\documentclass[10pt,letterpaper]{article}

\usepackage{cogsci}

\cogscifinalcopy % Uncomment this line for the final submission 

\usepackage{pslatex}
\usepackage{apacite}
\usepackage{float} % Roger Levy added this and changed figure/table
\usepackage{graphicx}
\usepackage{mathtools}
\usepackage{CJKutf8}
\usepackage{easybmat}
\usepackage{amsfonts}

\title{Compositional Neural Machine Translation by Removing the Lexicon from Syntax\\ COGSCI DRAFT SUMBISSION}
 
\author{{\large \bf Tristan Thrush (tristant@mit.edu)} \\
  MIT Department of Brain and Cognitive Sciences, 43 Vassar Street \\
  Cambridge, MA 02139 USA}

\begin{document}

\maketitle

\begin{abstract}
The meaning of a natural language utterance is largely determined from its syntax and words. Additionally, there is evidence that humans process an utterance by separating knowledge about the lexicon from syntax knowledge. Theories from semantics and neuroscience claim that complete word meanings are not encoded in the representation of syntax. In this paper, we propose neural units that can enforce this constraint over an LSTM encoder and decoder. We demonstrate that our model achieves competitive performance across a variety of domains including semantic parsing, syntactic parsing, and English to Mandarin Chinese translation. In these cases, our model outperforms the standard LSTM encoder and decoder architecture on many or all of our metrics. To demonstrate that our model achieves the desired separation between the lexicon and syntax, we analyze its weights and explore its behavior when different neural modules are damaged. When damaged, we find that the model displays the knowledge distortions that aphasics are evidenced to have.
\footnote{The code, trained models, tokenized data, and supplemental information about the training, can be found at: \textit{NOT INCLUDED IN DRAFT SUBMISSION}}

\end{abstract}

\section{Introduction}

Studies of Broca's and Wernicke's aphasia provide evidence that our brains understand an utterance by creating separate representations for word meaning and word arrangement \cite{aphasia}. There is a related thesis about human language, present across many theories of semantics, which is that syntactic categories are partially agnostic to the identity of words \cite{semantics}. This regularity in how humans derive meaning from an utterance is applicable to the task of natural language translation. This is because, by definition, translation necessitates the creation of a meaning representation for an input. According to the cognitive and neural imperative, we introduce new units to regularize an artificial neural encoder and decoder \cite{machine-translation}. These are called the Lexicon and Lexicon-Adversary units (collectively, LLA). Tests are done on a diagnostic task, and naturalistic tasks including semantic parsing, syntactic parsing, and English to Mandarin Chinese translation. We evaluate a Long Short-Term Memory (LSTM) \cite{lstm} encoder and decoder, with and without the LLA units, and show that the LLA version achieves superior translation performance. In addition, we examine our model's weights, and its performance when some of its neurons are damaged. We find that the model exhibits the knowledge and the lack thereof expected of a Broca's aphasic \cite{aphasia} when one module's weights are corrupted. It also exhibits that expected of a Wernicke's aphasic \cite{aphasia} when another module's weights are corrupted.

\section{Neural Motivation}

\citeauthor{aphasia} \citeyear{aphasia} showed that Broca's aphasics were able to understand ``the apple that the boy is eating is red'' with significantly higher accuracy than ``the cow that the monkey is scaring is yellow,'' along with similar pairs. The critical difference between these sentences is that, due to semantic constraints from the words, the first can be understood if it is presented as a set of words. The second cannot. This experiment provides evidence that the rest of the language neurons in the brain (largely Wernicke's area) can yield an understanding of word meanings but not how words are arranged. This also suggests that Broca's area builds a representation of the syntax.

In the same study, Wernicke's aphasics performed poorly regardless of the sentence type. This provides evidence that Broca's area cannot yield an understanding of word meanings.

Taken together, the two experiments support the theory that Broca's area creates a representation of the syntax without encoding complete word meanings. These other lexical aspects are represented separately in Wernicke's area, which does not encode syntax.

\section{Cognitive Motivation}

A tenet of generative grammar theories is that different words can share the same syntactic category \cite{semantics}. It is possible, for example, to know that the syntax for an utterance is a noun phrase that is composed of a determiner and a noun, followed by a verb phrase that is composed of a verb. One can know this without knowing the words. This also means that there are aspects of a word's meaning that the syntax does not determine; by definition, these aspects are invariant to word arrangement.

\section{Model}

In a natural language translation setting, suppose that an input word corresponds to a set of output tokens independently of its context. Even though this information might be useful to determine the syntax of the input utterance in the first place, the syntax does not determine this knowledge at all (by supposition). So, we can impose the constraint that our model's representation of the input's syntax cannot contain this context-invariant information. This regularization is strictly preferable to allowing all aspects of word meaning to propagate into the input's syntax representation.\footnote{While not disadvantaging our model over the standard one, our implementation choice of looking for lexical invariants in the identity of the output tokens may not always provide significant improvements. Consider the task of translating English into a binary code.} Without such a constraint, all inputs could, in principle, be given their own syntactic categories. This scenario is refuted by cognitive and neural theories. We incorporate the regularization with neural units that can separate representations of word meaning and arrangement.

With the exception of the equations that we list below, the encoding and decoding follows standard paradigms \cite{machine-translation}. The input at a time step to the LSTM encoder is a vector embedding for the input token. The final hidden and cell states of the encoder are the starting hidden and cell states of the LSTM decoder. The decoder does not take tokens as inputs; it decodes by relying solely on its hidden and cell states. The $t$th output, $o_t$, from the decoder is Softmax$(W(h_t))$, where $W$ is a fully connected layer and $h_t$ is the decoder's $t$th hidden state. $o_t$ is the length of the output dictionary. $o_t$'s index with the highest value corresponds to the token choice. The encoder and decoder's weights are optimized with the negative log likelihood loss. The inputs to the loss function are the log of the model's output and the ground-truth at each time step. Below, we describe our modifications.
\begin{align*}
l &= \sigma (\vee(w_1, w_2,  \cdots, w_m))\\
l_a &= \sigma (W_{a_2}(ReLU(W_{a_1}(\mbox{GradReverse}(h_e^\frown c_e)))))\\
o'_t &= l \odot o_t
\end{align*}
\noindent
Where:

$m$ is the number of input tokens.

$w_i$ is a vector embedding for the $i$th input token, and is the length of the output dictionary. It is not the same embedding used by the encoder LSTM.

$\sigma$ is the Sigmoid function.

$\vee$ is the max pooling of a sequence of vectors of the same length. The weight at the output vector's $i$th index is the max of all input vectors' weights at their $i$th indices.

$h_e$ and $c_e$ are the final hidden and cell states of the encoder.

$W_{a_1}$ and $W_{a_2}$ are fully connected layers.

$\frown$ is concatenation.

$\odot$ is the elementwise product.

GradReverse multiplies the gradient by a negative number upon backpropagation.
\\

$l$ is the output of the Lexicon Unit. Due to the max pooling, only one input token can be responsible for the value at a particular index of the output vector. The weights, $w_i$, are optimized solely by computing the binary cross entropy (BCE) loss between $l$ and the indicator vector where the $k$th element is 1 if the $k$th token in the output dictionary is in the output and 0 otherwise. This procedure forces a $w_i$ to represent the output tokens that are associated with its respective input token, without relying on aggregated contributions from the presence of several input tokens, and independently of the input word order.

$l_a$ is the output of the Lexicon-Adversary Unit. Its weights are optimized according to the BCE loss with $l$ as the target. This means that $l_a$ is the Lexicon-Adversary Unit's approximation of $l$. Because $h_e$ and $c_e$ are passed through a gradient reversal layer, the LSTM encoder is regularized to produce a representation that does not include information from $l$. Consequently, the LSTM decoder does not have this information either.

$o'_t$ is the $t$th output of our model. It can be converted to a token by finding the index with the highest weight. It is the result of combining $l$ via an elementwise product with the information from the regularized decoder.

\begin{figure}[ht]
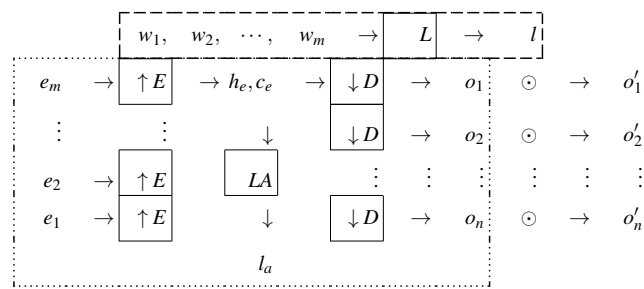

\begin{center}
\resizebox{\columnwidth}{!}{
$
\begin{BMAT}(b){rrrrrrrrrrrr}{cccccc}
&& w_1, & w_2, & \cdots, & w_m & \rightarrow & L & \rightarrow & l &&\\
e_m & \rightarrow & \uparrow E & \rightarrow & h_e, c_e & \rightarrow & \downarrow D & \rightarrow & o_1 & \odot & \rightarrow & o'_1\\ 
\vdots && \vdots && \downarrow && \downarrow D & \rightarrow & o_2 & \odot & \rightarrow & o'_2\\ 
e_2 & \rightarrow & \uparrow E && LA && \vdots & \vdots & \vdots & \vdots & \vdots & \vdots\\ 
e_1 & \rightarrow & \uparrow E && \downarrow && \downarrow D & \rightarrow & o_n & \odot & \rightarrow &  o'_n\\
&&&& l_a &&&&&&&
\addpath{(7,5,0)ruld}
\addpath{(4,2,0)ruld}
\addpath{(2,4,0)ruld}
\addpath{(2,2,0)ruld}
\addpath{(2,1,0)ruld}
\addpath{(6,4,0)ruld}
\addpath{(6,3,0)ruld}
\addpath{(6,1,0)ruld}
\addpath{(0,0,.)rrrrrrrrruuuuulllllllllddddd}
\addpath{(2,5,:)rrrrrrrrulllllllld}
\end{BMAT}
$
}
\end{center}
\caption{A graphic of our model. In addition to the terms described under our equations, we depict $e$ terms which are embeddings for input tokens, for use by the LSTM encoder. The LSTM encoder is $E$, the LSTM decoder is $D$, the Lexical Unit is $L$ and the Lexicon-Adversary Unit is $LA$. The dotted area contains the representation for the input's syntax, adversarially regularized to not include context-invariant lexical information, $l$. The dashed area contains the representation for this lexical information, which does not have syntax knowledge. At every output step, $l$ is combined with the decoder's output via an elementwise product.} 
\label{model}
\end{figure}

The recurrent encoder and decoder are the only modules that can represent the syntax, but they lack the expressivity to encode all potential aspects of word meaning. So, they are not always capable of producing a theoretically denied representation by giving all words their own syntactic category. The Lexicon Unit can represent these missing lexical aspects, but it lacks the expressivity to represent the syntax. See Figure \ref{model} for the model.

\section{Experiments}

We used \citeauthor{lake}'s \citeyear{lake} small diagnostic, the Geoquery semantic parsing dataset \cite{geoquery}, the Wall Street Journal syntactic parsing dataset of sentences up to length 10 \cite{wsj}, and the Tatoeba \cite{tatoeba} English to Chinese translation dataset processed by \citeauthor{tatoeba-data} \citeyear{tatoeba-data}.

To avoid the biases that can be introduced with hyperparameter tuning, we used the same hyperparameters with every model on every domain. These were chosen arbitrarily and kept after they enabled all models to reach a similar train accuracy (typically, close to 100 percent) and after they enabled all models to achieve a peak validation performance and then gradually yield worse validation scores. The hyperparameters are as follows. LSTM hidden size = 300, Lexicon Unit batch size = 1, batch size for other modules = 30, epoch to stop training the Lexicon Unit and start training other modules = 30, epoch to stop training = 1000, and Lexicon-Adversary Unit hidden size = 1000. The optimizer used for the Lexicon Unit was a sparse implementation of Adam \cite{adam} with a learning rate of 0.1 and otherwise the default PyTorch settings \cite{pytorch-optim}. In the other cases it was Adam \cite{adam} with the default PyTorch settings \cite{pytorch-optim}. The gradient through the encoder from the adversary's gradient reversal layer is multiplied by -0.0001. Additionally, the validation score is calculated after each train epoch and the model with the best is tested. To compute the Lexicon Unit to use, we measure its loss (BCE) on the validation set. Unless otherwise stated, we use the mean number of exact matches as the validation metric for the full model.

To judge overall translation performance, we compared the LLA-LSTM encoder and decoder with the standard LSTM encoder and decoder. We also compared our model with one that does not have the adversary but is otherwise identical. The LLA-LSTM model shows improvements over the standard model on many or all of the metrics for every naturalistic domain. Many of the improvements over the other models are several percentage points. In the few scenarios where the LLA-LSTM model does not improve upon the standard model, the discrepancy between the models is small. The discrepancy is also small when the LLA-LSTM model with no adversary performs better than the LLA-LSTM model. Table \ref{results} displays the test results across the domains.

\begin{figure*}[ht]
\begin{center}
\resizebox{\textwidth}{!}{
\begin{tabular}{cccc}
\includegraphics[scale=0.4]{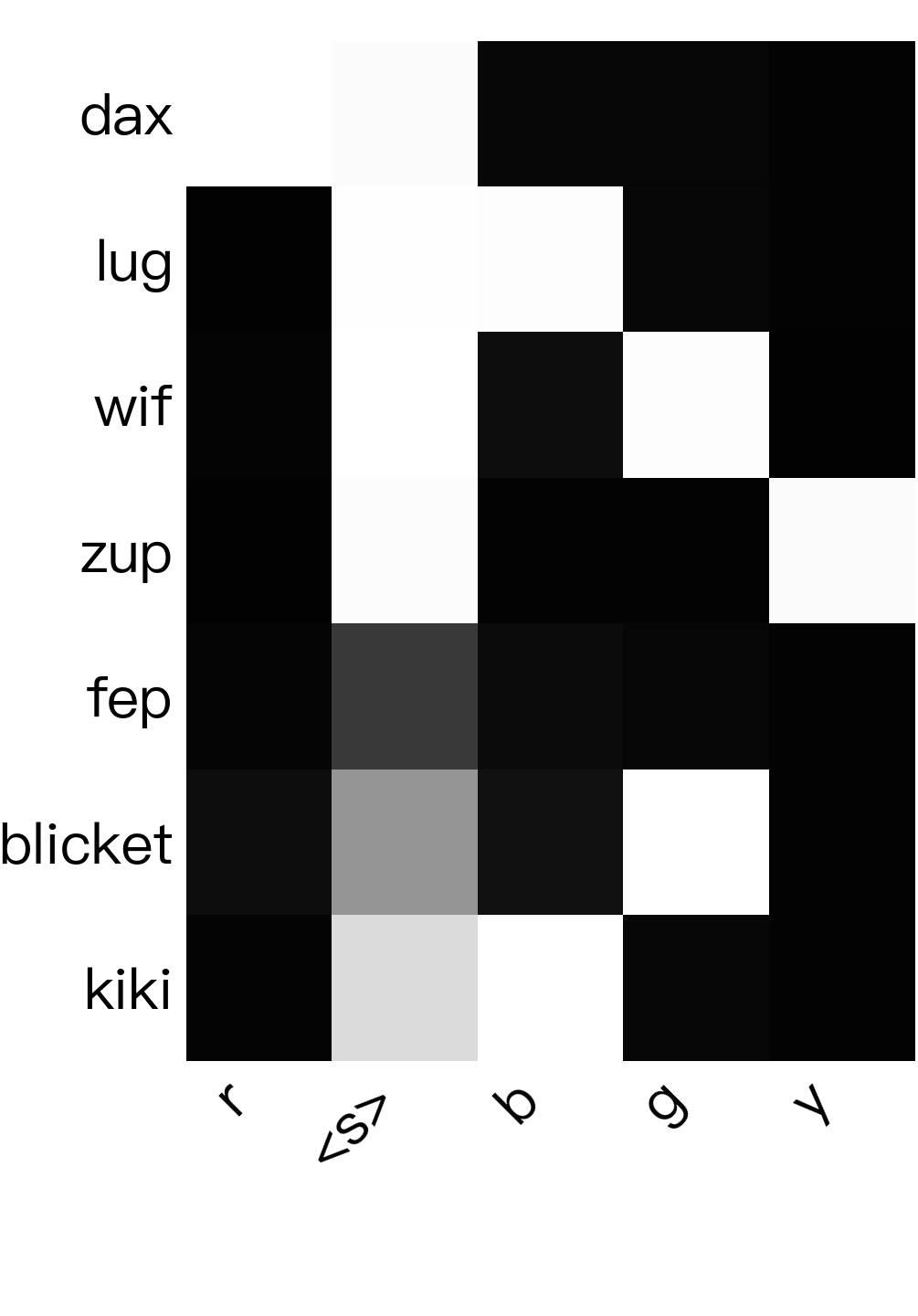} & \includegraphics[scale=0.4]{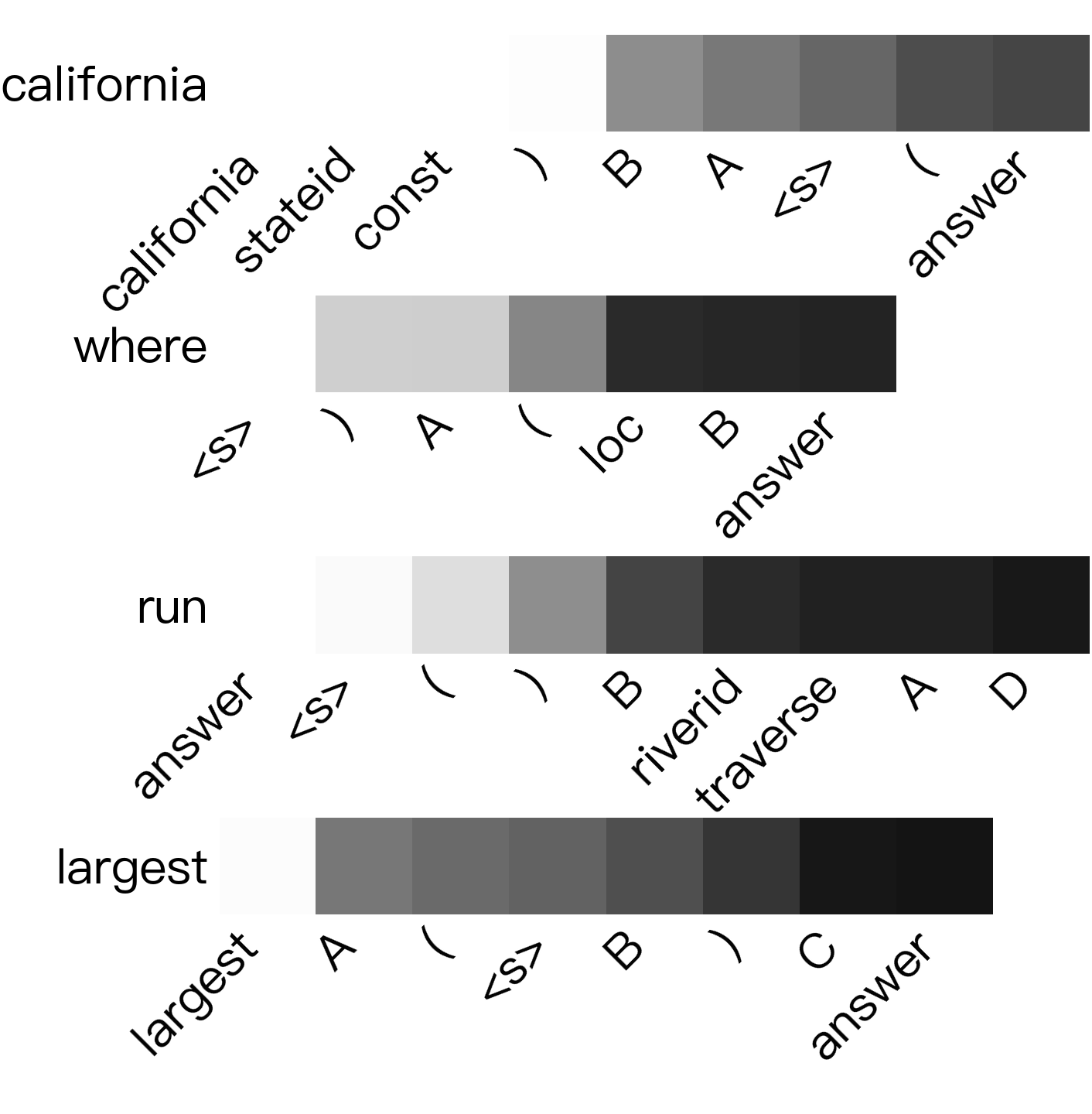} & \includegraphics[scale=0.4]{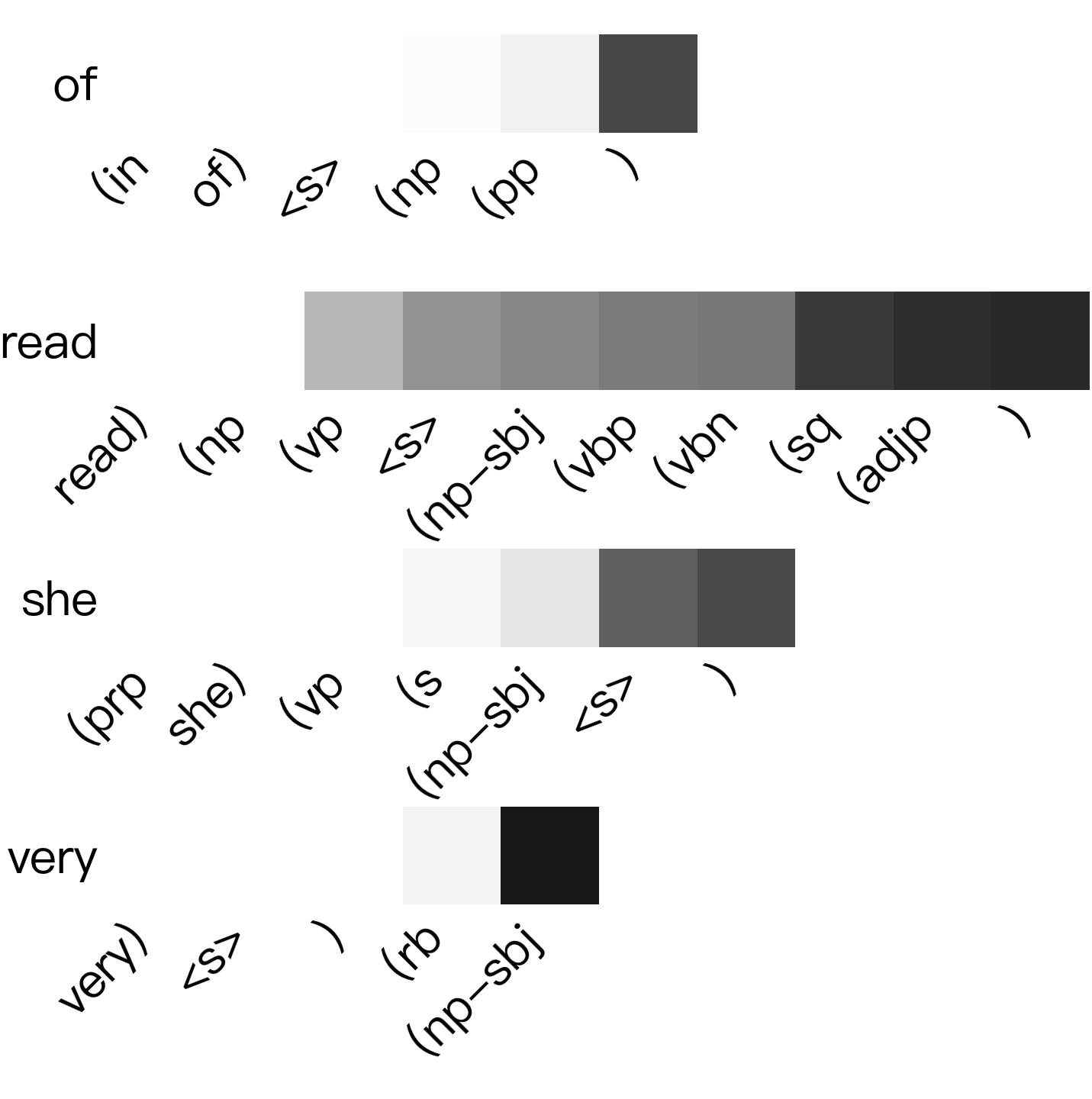} & \includegraphics[scale=0.4]{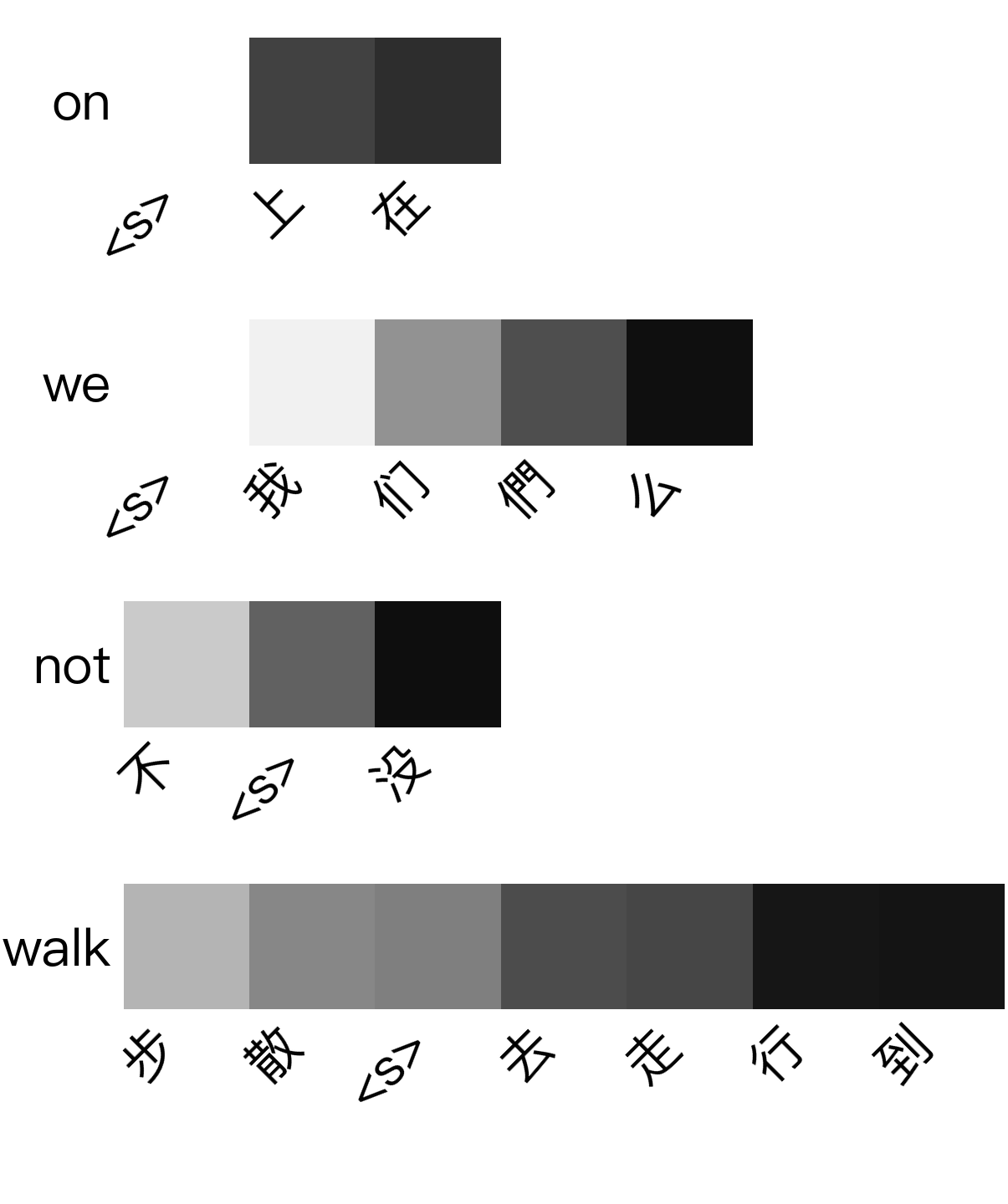}\\
(a) & (b) & (c) & (d)\\
\end{tabular}
}
\end{center}
\caption{The learned $\sigma (w)$'s for some diverse input words; they are arbitrarily chosen, subject to the constraints listed in the text. Black is zero and white is one. (a) shows the results from the colors dataset; the input and output dictionaries are so small that all of the weights for all of the input words are shown. (b), (c), and (d) show results from the semantic parsing, syntactic parsing, and English to Mandarin translation datasets, respectively. Because there are so many words in the input dictionary, only four $\sigma (w)$'s are shown in each case. Because there are so many tokens in the output dictionary, if a weight within a $\sigma (w)$ is zero when rounded to the nearest tenth's place, then it is omitted. So, for each $\sigma (w)$, approximately 170, 9480, and 3430 weights were omitted for the semantic parsing, syntactic parsing, and English to Mandarin translation cases, respectively.} 
\label{heatmaps}
\end{figure*}

Additionally, we provide evidence that the model learns knowledge of a separation between syntax and the lexicon that is similar to that of a human. Figure \ref{heatmaps} displays the learned $\sigma (w)$ embeddings for some input words, across the domains. To avoid cherry-picking the results, we chose the input words arbitrarily, subject to the following constraint. We considered each word to typically have a different syntactic category than the other choices from that domain. This constraint was used to present a diverse selection of words. Table \ref{aphasiaresults} displays the output behavior of models that we damaged to resemble the damage that causes aphasia. To avoid cherry-picking the results, we arbitrarily chose an input for each domain, subject to the following constraint. The input is not in the train set and the undamaged LLA-LSTM model produces a translation that we judge to be correct. For all inputs that we chose, damage to the analog of Broca's area (the LSTMs) results in an output that describes content only if it is described by the input. However, the output does not show understanding of the input's syntax. In the naturalistic domains, damage to the analog of Wernicke's area (the Lexicon Unit) results in an output with incorrect content that would be acceptable if the input had different words but the same syntax. These knowledge distortions are precisely those that are expected in the respective human aphasics \cite{aphasia}. We also provide corpus-level results from the damaged models by presenting mean precision on the test sets. Because the output languages in all of our domains use tokens to represent meanings in many cases, it is expected that the analog to Wernicke's area is responsible for maintaining a high precision.

\begin{table*}
\begin{center}
\caption{Comparison of the models on the test sets. The metrics used are mean precision (Prec.), mean recall (Rec.), mean accuracy (Acc.), mean number of exact matches (Exact.), and corpus-level BLEU \cite{bleu}.}
\label{results}
\vskip 0.12in
\resizebox{\textwidth}{!}{
\begin{tabular}{llllllllllllll}
\hline
                     & \multicolumn{4}{l}{Colors}  & \multicolumn{4}{l}{GEO} & \multicolumn{4}{l}{WSJ10} &
                     \multicolumn{1}{l}{English to Mandarin}\\
Encoder and Decoder Model                & Prec. & Rec. & Acc. & Exact & Prec. & Rec. & Acc. & Exact & Prec.  & Rec.  & Acc.  & Exact & BLEU \\ \hline
LLA-LSTM & 90 & 62.50 & \textbf{50.90} & 0 & 97.00 & 92.65 & 88.31 & \textbf{50.4} & \textbf{76.81} & 56.90 & 42.49 & 2.51 & \textbf{11.71}\\
LLA-LSTM (No Adversary) & 90 & 62.50 & 49.48 & 0 & \textbf{97.08} & \textbf{92.68} & \textbf{89.80} & 47.2 & 73.42 & \textbf{57.85} & 43.67 & 2.76 & 11.40\\
LSTM & 90 & 62.50 & 49.48 & 0 & 93.66 & 91.18 & 88.60 & 37.6 & 65.94 & 56.74 & \textbf{44.48} & 2.76 & 9.95\\ \hline
\end{tabular}
}
\end{center}
\end{table*}

\begin{table*}
\begin{center}
\caption{Results for artificial Wernicke's and Broca's aphasia induced in the LLA-LSTM model. Damage to neural modules is done by randomly initializing their weights, causing the loss of all learned information. The inputs that we present are arbitrarily chosen, subject to the constraints listed in the text. Mean precision (Prec.) results on the test sets are also provided to demonstrate corpus-level results. An ellipses represents the repetition of the preceding word at least 1000 times.}
\label{aphasiaresults}
\vskip 0.12in
\resizebox{\textwidth}{!}{
\begin{tabular}{lllll}
\hline
& \multicolumn{2}{l}{Colors}  & \multicolumn{2}{l}{GEO}\\
Modules Damaged & wif kiki lug & Prec. & what is the capital of utah & Prec \\ \hline
None & b g & & answer ( A ( capital ( A ) loc ( A B ) const ( B stateid ( utah ) ) ) ) ) &\\
LSTMs & $<s>$ & \textbf{95} & ( ( const ... & \textbf{99.09}\\
Lexicon Unit & b g & 90 & answer ( A ( capital ( A ) loc ( A B ) const ( B stateid ( michigan ) ) ) ) ) & 93.96\\ \hline
& &\\
\hline
& \multicolumn{2}{l}{English to Mandarin}  & \multicolumn{2}{l}{WSJ10}\\
Modules Damaged & I ate some fish . & Prec. & he needs it & Prec. \\ \hline
None & \begin{CJK*}{UTF8}{bsmi} 我 吃 了 一 些 魚 。\end{CJK*} & & (s (np-sbj (prp he) ) (vp (vbz needs) (np (prp it) ) ) ) &\\
LSTMs & \begin{CJK*}{UTF8}{bsmi} 吃 我 \end{CJK*} ... & \textbf{87.37} & (np ... & \textbf{74.12}\\
Lexicon Unit & \begin{CJK*}{UTF8}{bsmi} 我 拿 了 這 條 話 。\end{CJK*} & 54.52 & (s (np-sbj (prp he) ) (vp (vbz will) (np (prp it) ) ) ) & 65.76\\ \hline
\end{tabular}
}
\end{center}
\end{table*}

\subsection{Sequences of Color}

The first experiment by \citeauthor{lake} \citeyear{lake} included a dataset of 14 training pairs and 10 test pairs. In the dataset, an input is a sequence of words from an artificial language created by the authors. An output is a sequence of colored dots. Because the dataset is so small, we use the train set as the validation set. The input and output dictionary are 7 and 4 words, respectively (not including the stop, ``$<s>$,'' token). In their paper, the authors argue that it is clear that the words have meanings. Four of the words correspond to unique output tokens, and three of them correspond to functions of the output tokens (for example, repeating the same dot three times). The dataset showcases the contrast between human and standard neural network responses. Their paper shows that humans had high accuracy on the test set, whereas standard neural models scored essentially zero exact matches.

The LLA-LSTM model that we tested appears to achieve only insignificantly higher results in Table \ref{results}. However, it has learned, from just 14 training examples, how to map some of the words to \citeauthor{lake}'s interpretation of their context-invariant meanings. This is shown in Figure \ref{heatmaps} (a). In the figure, ``dax,'' ``lug,'' ``wif,'' and ``zup'' are interpreted correctly to mean ``r,'' ``g,'' ``b,'' and ``y,'' respectively. Here, the letters correspond to the types of unique dots, which are red, green, blue, and yellow, respectively. The other words, ``fep,'' ``kiki,'' and ``blicket,'' are taken by \citeauthor{lake} to have functional meanings, and so are correctly not associated strongly with any of the output tokens. The exceptions are two erroneous associations between ``kiki'' and blue and ``blicket'' and green. Also, every sentence has a stop token, so the LLA units learned that the context-invariant meanings of each word include it. The LLA units can handle cases where a word corresponds to multiple output tokens, and the output tokens need not be monolithic in the output sequence. As shown in tests from all of the other domains, these output token correspondences may or may not be relevant depending on the specific context of a word, but the recurrent component of the architecture is capable of determining which to use.

\subsection{Semantic Parsing}

Geoquery (GEO) is a dataset where an input is an English geography query and the corresponding output is a parse that a computer could use to look up the answer in a database \cite{geoquery}. We used the standard test set of 250 pairs from \citeauthor{geoquery-web} \citeyear{geoquery-web}. The remaining data were randomly split into a validation set of 100 pairs and a train set of 539 pairs. We tokenized the input data by splitting on the words and removing punctuation. We tokenized the output data by removing commas and splitting on words, parentheses, and variables. There are 283 tokens in the input dictionary and 177 tokens in the output dictionary, respectively.

Figure \ref{heatmaps} (b) shows some weights for four input words, which are all relevant to the inputs. Many of the weights correspond directly to the correct predicates. Other tokens have high weights because they are typically important to any parse. These are parentheses, variables (A, B, C, and D), the ``answer'' token, and the stop token.

\subsection{Syntactic Parsing}

The Wall Street Journal portion of the Penn Treebank is a dataset where English sentences from The Wall Street Journal are paired with human-generated phrase parses \cite{wsj}. We use the test, validation, and train set from \citeauthor{kim}'s \citeyear{kim} paper. For efficiency, we only use sentences that have 10 or fewer words, lowercase all words, and modify \citeauthor{kim}'s output data so that left parentheses are paired with their corresponding nonterminal and right parentheses are paired with their corresponding terminal. The input and output data were both tokenized by splitting where there is a space. The test, validation, and train set are 398, 258, and 6007 pairs, respectively. There are 9243 tokens in the input dictionary and 9486 tokens in the output dictionary.

Figure \ref{heatmaps} (c) shows some weights for four input words. They all highlight the relevant terminal, and syntactic categories that are usually associated with that word. The associated categories typically are either those of that word, the phrases headed by the category of that word, or those that select or are selected by that word. The relevant nonterminal terminology is as follows \cite{wsj}: ``(in'' is a preposition or subordinating conjunction, ``(np'' is a noun phrase, ``(pp'' is a prepositional phrase, ``(np-subj'' is a noun phrase with a surface subject marking, ``(vp'' is a verb phrase, ``(vbn'' is a verb in the past participle, ``(adjp'' is an adjective phrase, ``(vbp'' is a non-3rd person singular present verb, ``(prp'' is a personal pronoun, ``(rb'' is an adverb, ``(sq'' is the main clause of a wh-question, or it indicates an inverted yes or no question, and ``(s'' is the root.

\subsection{English to Chinese}

The Tatoeba \cite{tatoeba} English to Chinese translation dataset, processed by \citeauthor{tatoeba-data} \citeyear{tatoeba-data}, is a product of a crowdsourced effort to translate sentences of a user's choice into another language. The data were split randomly into a test, validation, and train set of 1500, 1500, and 18205 pairs, respectively. The English was tokenized by splitting on punctuation and words. The Chinese was tokenized by splitting on punctuation and characters. There are 6919 and 3434 tokens in the input and output dictionary, respectively. There are often many acceptable outputs when translating one natural language to another. As a result, we use the corpus-level BLEU score \cite{bleu} to test models and score them on the validation set.

Figure \ref{heatmaps} (d) shows some weights for four input words. The listed Chinese words are an acceptable translation (depending on the context) and correspond roughly one-to-one with the English inputs. There are three exceptions. Although \begin{CJK*}{UTF8}{bsmi} 么 \end{CJK*} is correctly given a low weight, its presence seems to be an error; it usually appears with another character to mean ``what.'' \begin{CJK*}{UTF8}{bsmi} 我 們 \end{CJK*} and \begin{CJK*}{UTF8}{gbsn} 我 们 \end{CJK*} typically translate to ``we,'' even though \begin{CJK*}{UTF8}{gbsn} 我 \end{CJK*} alone translates to ``me.'' \begin{CJK*}{UTF8}{bsmi} 們 \end{CJK*} is a plural marker and \begin{CJK*}{UTF8}{gbsn} 们 \end{CJK*} is the same, but simplified; both versions evidently found their way into the dataset. The network has correctly learned to associate both Chinese words necessary to form the meaning of ``we.'' Also, \begin{CJK*}{UTF8}{gbsn} 步散 \end{CJK*} means ``walk,'' but \begin{CJK*}{UTF8}{gbsn} 散 \end{CJK*} generally does not appear alone to mean ``walk.'' Again, the network has learned to correctly associate all of the necessary characters with an input word. 

The results from this dataset in Table \ref{aphasiaresults} warrant a discussion for readers who do not know Chinese. As in the other cases, the model demonstrates the expected knowledge and lack thereof when different types of artificial aphasia are induced.\footnote{We consulted with a native Mandarin Chinese speaker to interpret the model's outputs.} The outputs are also productions that Chinese aphasics are expected to make per \citeauthor{aphasia}'s \citeyear{aphasia} description. When the model is undamaged, its output is a correct translation for ``I ate some fish.'' When the model's LSTMs are damaged (simulating the conditions for Broca's aphasia), the production has incorrect syntax, and translates word for word to ``eat I ...'' These are both correct content words. When the model's Lexicon Unit is damaged (simulating the conditions for Wernicke's aphasia), the production has correct syntax. Impressively, the Chinese actually has the same syntax as the correct translation for ``I ate some fish.'' However, the content is nonsensical. The English translation is ``I took the utterance.'' Compared to the correct Mandarin translation, this incorrect one has the same subject and the same past-tense marker, \begin{CJK*}{UTF8}{gbsn} 了 \end{CJK*}, for the verb. However it uses a different verb, object, and determiner.

\section{Related Work}

There is evidence that generic attention mechanisms for machine translation already utilize the thesis that words have meanings that are independent of syntax. They learn correspondences between output tokens and a hidden state produced immediately after an encoder reads a particular input word \cite{attention-original}. But the same mechanism is not at play in our model. Generic attention mechanisms do not necessarily impose a constraint on the input's syntax representation. Additionally, typical objective functions do not explicitly link input words with invariance in the output. Finally, one does not need to choose either LLA units or attention. LLA units can be incorporated into recurrent neural network systems with attention or other machine transduction architectures such as transformers \cite{transformer}.

Recent work has incorporated some of the ideas in our paper into a neural machine translation model with the use of a specific attention mechanism \cite{attention}. But the authors only demonstrate success on a single artificial dataset with a lexicon of about ten words, and they did not explore the effects of damaging parts of their model. Their optimization procedure also does not prohibit context-invariant lexical information from passing through the recurrent portion of their model. This incorrectly allows the possibility for a representation to be learned that gives every input word its own syntactic category. Lastly, their architecture provides a softer constraint than the one that we demonstrate, as information from several input words can aggregate and pass through the non-recurrent module that they use.

There are other attempts to incorporate theories about human language to regularize a transduction model, but many have not scaled to the level of generality that the LLA units and some attention architectures show. These include synchronous grammars \cite{synchronous}, data augmentation \cite{augmentation}, Meta learning \cite{meta}, and hard-coded maps or copying capabilities from input to output \cite{dependency} \cite{data-recombination}. All require hard-coded rules that are often broken by the real world.

\section{Conclusion}

Neural and cognitive theories provide an imperative for computational models to understand human language by separating representations of word meanings from those of syntax. Using this constraint, we introduced new neural units that can provide this separation for the purpose of translating human languages. When added to an LSTM encoder and decoder, our units showed improvements in all of our experiment domains over the typical model. The domains were a small artificial diagnostic dataset, semantic parsing, syntactic parsing, and English to Mandarin Chinese translation.  We also showed that the model learns a representation of human language that is similar to that of our brains. When damaged, the model displays the same knowledge distortions that aphasics do. 

\section{Acknowledgments}

\textit{NOT INCLUDED IN DRAFT SUBMISSION}

\bibliographystyle{apacite}

\setlength{\bibleftmargin}{.125in}
\setlength{\bibindent}{-\bibleftmargin}

\bibliography{CogSci_Template}

\end{document}